\documentclass{article}

\usepackage[final]{neurips_2024}

\usepackage[utf8]{inputenc} % allow utf-8 input
\usepackage[T1]{fontenc}    % use 8-bit T1 fonts
\usepackage{hyperref}       % hyperlinks
\usepackage{url}            % simple URL typesetting
\usepackage{booktabs}       % professional-quality tables
\usepackage{amsfonts}       % blackboard math symbols
\usepackage{nicefrac}       % compact symbols for 1/2, etc.
\usepackage{microtype}      % microtypography
\usepackage{xcolor}         % colors
\usepackage{multicol}
\usepackage{multirow}
\usepackage{tabularray}
\usepackage{amsmath}
\usepackage{listings}
\usepackage{graphicx}
\usepackage{caption}  % Optional: For captions beside code blocks

\usepackage{svg}

\title{LLM Vocabulary Compression for Low-Compute Environments}

\author{%
  Sreeram Vennam \\
  IIIT Hyderabad \\
  \texttt{sreeram.vennam@students.iiit.ac.in} \\
  \And
  Anish Joishy \\
  IIIT Hyderabad \\
  \texttt{anish.joishy@research.iiit.ac.in} \\
  \AND
  Ponnurangam Kumaraguru \\
  IIIT Hyderabad \\
  \texttt{pk.guru@iiit.ac.in} \\
}

\begin{document}

\maketitle

\begin{abstract}
    We present a method to compress the final linear layer of language models, reducing memory usage by up to 3.4x without significant performance loss. By grouping tokens based on Byte Pair Encoding (BPE) merges, we prevent materialisation of the memory-intensive logits tensor. Evaluations on the TinyStories dataset show that our method performs on par with GPT-Neo and GPT2 while significantly improving throughput by up to 3x, making it suitable for low-compute environments.
\end{abstract}

\section{Introduction}

The global trend in machine learning is increasingly focused on scaling with larger numbers of GPUs \cite{musk2024gpumegacluster, nextplatform2024gpu}. In contrast, many researchers operate in low-compute environments, a disparity often referred to as the \textit{compute divide} \cite{besiroglu2024computedividemachinelearning}. Our work seeks to address this gap by optimising the utilisation of compute-constrained environments. Specifically, we target the vocabulary layer in language models. During training or fine-tuning, it becomes necessary to materialise a tensor of shape \texttt{[batch\_size, sequence\_length, vocab\_size]}\label{logits-tensor}. Even with conservative values for the batch size and sequence length, such as a tensor of shape \texttt{[32, 512, 50000]}, this alone consumes approximately $3.32$ GB of memory. For the remainder of this text, we will refer to this tensor as the \textit{logits tensor}.

Although no research has directly addressed this specific tensor, several studies have acknowledged the vocabulary layer as a computational bottleneck and have proposed methods to reduce its computational complexity \cite{jozefowicz2016exploringlimitslanguagemodeling}. One of the earliest efforts was by \cite{goodman2001classes}, who introduced a class-based approach by organising tokens into classes and training two separate models: one for predicting the class and another for predicting the actual token. Later, \cite{joulin2017efficient} proposed a hierarchical organisation of tokens based on their frequency, although computing these frequencies remains computationally expensive. Building on this, \cite{baevski2018adaptive} extended the hierarchical approach to allow for variable capacity inputs. However, a comprehensive analysis of memory usage in this context is still lacking.

In this work, we propose a method to reduce the memory footprint of the final embedding layer by grouping tokens and predicting the final token in a two-step process effectively compressing the vocabulary layer. Our approach differentiates from the work of \cite{joulin2017efficient, goodman2001classes} through two key innovations. First, rather than grouping tokens based on their frequency—a process that necessitates significant pre-processing—we group tokens according to the ordering of BPE \cite{sennrich-etal-2016-neural} merges, which inherently exploits token occurrence patterns. Second, we observe that it is unnecessary to use two separate models to predict the group and the token. Instead, simple linear layers over the hidden states can simultaneously learn both the group and token predictors. We empirically demonstrate that our modification does not negatively impact model performance.

\section{Methodology}\label{method}

\subsection{Problem Formulation}

Our goal is to produce a mapping from the hidden state \( h \) to a probability distribution over the entire vocabulary. The resulting probability distribution is used with \texttt{top\_k}, sampling and other techniques during decoding.

\[
P(v \mid h) = \text{Softmax}(W h + b)
\]

Where \( P(v \mid h) \) is the probability distribution over the vocabulary given the hidden state \( h \), \( W \) is the large weight matrix, and \( b \) is the bias vector.

Producing this distribution for every single token leads to materialising the logits tensor, therefore, to avoid this, during training, we want to materialise only a subset of this distribution, and during inference, we only need to produce this distribution for the last hidden state which is used to predict the next token in combination with techniques such as \texttt{top\_k} and sampling.

% \begin{figure}[htbp]
%     \centering
%     \rule{10cm}{4cm}  % Placeholder for a figure with width 6cm and height 4cm
%     \caption{Sample placeholder for a figure with specified size.}
%     \label{fig:placeholder}
% \end{figure}

\subsection{Definitions and Notation}

\paragraph{Grouping}{
First, we divide the vocabulary into \(G\) groups based on their token indices. Each group contains tokens from consecutive index ranges. Let $S_g$ and $E_g$ be the starting index and ending index respectively for tokens part of group $g = 0,1,\dots,G-1$.

\begin{align*}
S_g &= \left\lfloor \frac{|v| \cdot g}{G} \right\rfloor, & 
E_g &= \left\lfloor \frac{|v| \cdot (g + 1)}{G} \right\rfloor - 1
\end{align*}

For example, if we only had two groups; group $0$ contains tokens from $0$ to $\lfloor |v| / 2 \rfloor - 1$ and group $1$ contains tokens from $\lfloor |v| / 2 \rfloor$ to $|v| - 1$. This partitioning implicitly makes use of BPE merging order, which is representative of the frequency of tokens.
}

\paragraph{Grouping Tensor}{
We define weight block $W_g \in \mathbb{R}^{d \times G}$ where $d$ is the hidden dimension of the transformer. This tensor is used to predict the group index from a hidden state.
}

\paragraph{Scale and Shift Tensors}{
We define a 3 weight blocks, $W_s \in \mathbb{R}^{d \times S}$, $W_{Pg} \in \mathbb{R}^{S}$, and $W_{Qg} \in \mathbb{R}^{S}$ where $S$ is the number of tokens per group and $g = 0,1,\dots,G-1$.

$W_s \in \mathbb{R}^{d \times S}$ it the shared linear tensor that is applied for all groups. $W_{Pg} \in \mathbb{R}^{S}$ is the scaling linear tensor specific to the group $g$. $W_{Qg} \in \mathbb{R}^{S}$ is the shifting linear tensor specific to the group $g$. These tensors are used to predict the exact token from the hidden state, once the group has been identified. We detail the operation in Section~\ref{applying-linears}.
}

\subsection{Method}

Our method works differently during training vs inference since these modes of operation have different requirements. Training requires extensive parallelism but our knowledge of the labels tell us what group to choose. Inference requires the complete probability distribution over the entire vocabulary, but it requires it only for the last hidden state.

\subsubsection{Training}

% During training, the model simultaneously learns to predict the correct group and the token within that group. To take the loss for tokens across all groups simultaneously, we normalize the tokens to the range $[0, g)$ which we call \texttt{token\_labels} The loss function is defined as:

During training, since we know the labels apriori, we already know what group each hidden state belongs to. Therefore, we only need to materialise the distribution for a token within a group, which is of size $S \le v$.

We also train the grouping tensor during training, therefore, we have two objectives we are trying to minimise. The grouping tensor loss $\mathcal{L}_{\text{group}}$ coming from predicting which group each hidden state belongs to, and the token loss $\mathcal{L}_{\text{token}}$ coming from predicting tokens within a group.

Assume $h \in \mathbb{R}^{d}$ is a single hidden state of dimension $d$. We know that this hidden state belongs to group $g$ and token $t$ within group $g$. First, we apply the grouping tensor on $h$ and calculate $\mathcal{L}_{\text{group}}$. To calculate $\mathcal{L}_{\text{token}}$, we first apply the linear block shared across groups $W_s$ on $h$. We then apply the scale and shift tensors for group $g$ which becomes $W_{Pg} \circ (W_s \cdot h) + W_{Qg}$. This tensor is then used against the label token $t$ to produce $\mathcal{L}_{\text{token}}$. We refer to this scale and shift transformation as "applying linears"\label{applying-linears}. The final objective we minimise is the sum of both losses.

$$
\mathcal{L}_{\text{group}} = \text{CrossEntropy}(W_g \cdot h, g), 
$$

$$
\mathcal{L}_{\text{token}} = \text{CrossEntropy}(W_{Pg} \circ (W_s \cdot h) + W_{Qg}, t)
$$

$$
\mathcal{L} = \mathcal{L}_{\text{group}} + \mathcal{L}_{\text{token}},
$$

Note that this transformation is not linear due to grouping, it forms a deeper network. An implementation for "applying linears" can be found in the Appendix~\ref{app:psuedo}.

\subsubsection{Inference}

In the inference phase, the model generates the probability distribution over the entire vocabulary $v$ from the last hidden state. We construct this distribution by using the token probability conditioning on the group probability.

Assume $h \in \mathbb{R}^{d}$ is a single hidden state of dimension $d$. We know that this hidden state belongs to group $g$ and token $t$ within group $g$. First, we apply the grouping tensor and \texttt{Softmax} to obtain a distribution over the groups for $h$. Then, for each group, we compute the token probability distribution by "applying linears"~\ref{applying-linears} and \texttt{Softmax} on $h$. We then multiply the token probabilities with their respective group probability. Finally we concatenate these distributions and the result is used in next token prediction.

$$
\mathcal{P}_{\text{group}} \Rightarrow \text{Softmax}(W_g \cdot h, g), 
$$

$$
\mathcal{P}_{\text{token} | g} \Rightarrow \text{Softmax}(W_{Pg} \circ (W_s \cdot h) + W_{Qg}, t)
$$

$$
\mathcal{P}_{\text{vocab}} \Rightarrow \text{Concat}\left( \mathcal{P}_{\text{group}}[0] \cdot \mathcal{P}_{\text{token} | 0}(t), \mathcal{P}_{\text{group}}[1] \cdot \mathcal{P}_{\text{token} | 1}(t), \dots, \mathcal{P}_{\text{group}}[G-1] \cdot \mathcal{P}_{\text{token} | G-1}(t) \right)
$$

Where $\mathcal{P}_{\text{group}}$ refers the the probability distribution over the groups, $\mathcal{P}_{\text{token} | g}$ refers to the probability distribution over tokens within the group $g$, and $\mathcal{P}_{\text{vocab}}$ refers to the probability distribution over the entire vocabulary.

\subsection{Optimal Memory Configuration}\label{sec:optimality}

% Fast vs Slow Implementation + $S = g = \sqrt{v}$
From the above description of our method, instead of materialising the logits tensor, we materialise \texttt{[batch\_size, sequence\_length, group\_size]} and \texttt{[batch\_size, sequence\_length, num\_groups]}. Let $b$ be the batch size, $s$ be the sequence length, $S$ be the group size and $G$ be the number of groups. We wish to minimise the memory usage of the combined tensors $[b, s, S]$ and $[b, s, G]$ under the condition that $G = \frac{v}{S}$. It is clear that this minima is achieved when $S = G = \sqrt{v}$, that is, we have $\sqrt{v}$ groups each containing $\sqrt{v}$ tokens.

% The tensors we materialise instead of the logits tensor $[b, s, v]$ takes up memory $[b, s, 2\sqrt{v}]$ which is a significant amount of memory saved. Taking the example from the Introduction~\ref{logits-tensor}, for batch size $32$, sequence length $512$ and vocabulary size $50000$, the combined memory usage of the tensors is $0.0273$ GB --- over 120x times smaller.

%%%%%%%%%%%%%%%%%%%%%%%%%%%%%%%%%%%%%%%%%%%%%%%%%%%%%%%%%%%%

\section{Performance}

\subsection{Language Modelling}

We test our approach on language modelling a task infamous for being infeasible on low compute devices. We apply our method on the GPT-2 architecture and compare memory usage against the base GPT-2 model and the GPT-Neo model.

\paragraph{Dataset}{
We train on the TinyStories \cite{eldan2023tinystories} dataset, the standard for testing small language models.
}

\paragraph{Metrics \& Evaluation}{
We adopt the LLM evaluation metrics from TinyStories. This includes evaluating grammar, creativity, consistency, plot coherency and the estimated age of the writer.
}

\paragraph{Setting}{
We replicate the settings used in TinyStories, we use batch size $32$ with sequence length $512$. GPT-Neo uses a window size of $256$. We don't however use a trained tokenizer since the vocabulary size was relatively small and isn't much of a bottleneck.

We train 3 different sized models based on hidden size per architecture to extensively evaluate our approach. Every model uses 8 layers and 8 heads per layer. All models were trained for a single epoch over the entire dataset. This is because we had limited access to compute and training multiple models for several epochs would take months for the compute we had available.
% Loss curves can be found in the Appendix~\ref{tinystories-val-loss-curves}.
}

\begin{table}[h]
\centering
\caption{Comparing Model Performance on TinyStories}
\begin{tabular}{l|l|p{1.5cm}p{1.5cm}p{1.7cm}p{1.5cm}p{1.5cm}}
\toprule
\textbf{Model} & \textbf{Hidden Size} & \textbf{Grammar} & \textbf{Creativity} & \textbf{Consistency} & \textbf{Plot} & \textbf{Age} \\
\midrule
\multirow{3}{*}{GPT-2} & 128 & 3.8 & 4.22 & 3.7 & 2.95 & 4.55 \\
                         & 256 & 4.65 & 3.6 & 6.05 & 4.82 & 4.75 \\
                         & 512 & 5.82 & 3.9 & 7.15 & 5.75 & 4.9 \\
\midrule
\multirow{3}{*}{GPT-Neo} & 128 & 3.7 & 3.72 & 4.05 & 2.97 & 4.5 \\
                         & 256 & 4.77 & 4.1 & 5.85 & 4.57 & 4.65 \\
                         & 512 & 5.15 & 4.17 & 6.65 & 5.22 & 5.0 \\
\midrule
\multirow{3}{*}{Ours} & 128 & 3.53 & 3.62 & 4.5 & 3.25 & 4.55 \\
                      & 256 & 4.85 & 4.3 & 5.87 & 4.47 & 4.7 \\
                      & 512 & 5.15 & 4.025 & 6.7 & 5.475 & 4.96 \\
\bottomrule
\end{tabular}
\label{tab:model_performance_comparison}
\end{table}

\paragraph{Results}{
Table~\ref{tab:model_performance_comparison} contains the results for the models we trained. First we see that model performance does increase with model size validating our setup. Our approach performs on par with GPT-Neo and GPT-2. These results strongly suggest that our approach is able to compress the vocabulary layer without significant loss in performance.
}

\subsection{Multiclass Classification}

For robustness, we evaluate our approach on a task disconnected from language modelling, image classification. Here grouping is arbitrary, but even in such conditions, our approach works surprisingly well. The implementation details for this experiment including the architecture used can be found in the Appendix~\ref{app:multiclass}.

\paragraph{Dataset}{
We generate $100 \times 100$ images with different attributes and style, each such combination of attributes is a label. We produce a synthetic dataset since real world datasets don't often have a large number of labels. Our dataset has $184320$ unique labels. Some examples from the dataset are presented in Figure~\ref{fig:examples}. A comprehensive description of the dataset is present in the Appendix~\ref{synthetic-dataset}.
}

\begin{figure}[h]
    \centering
    \includegraphics[width=1\linewidth]{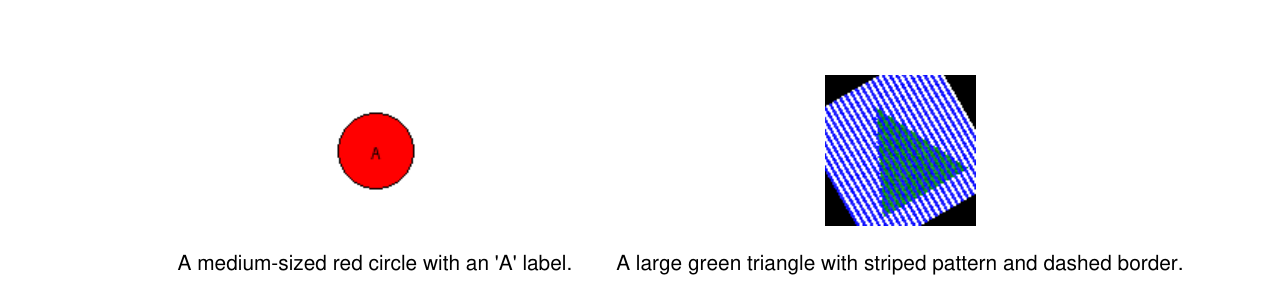}
    \caption{Example images from the synthetic dataset for image classification.}
    \label{fig:examples}
\end{figure}

\paragraph{Results}{
Figure~\ref{fig:synthetic-val-accuracy} plots validation accuracy across training steps. We find that our approach enables learning in scenarios in which learning would otherwise not have occurred despite using significantly fewer parameters. Our method improves performance since learning the group~\ref{app:groupaccuracyimg} is simpler.

\begin{figure}[h]
    \centering
    \includegraphics[width=0.7\linewidth]{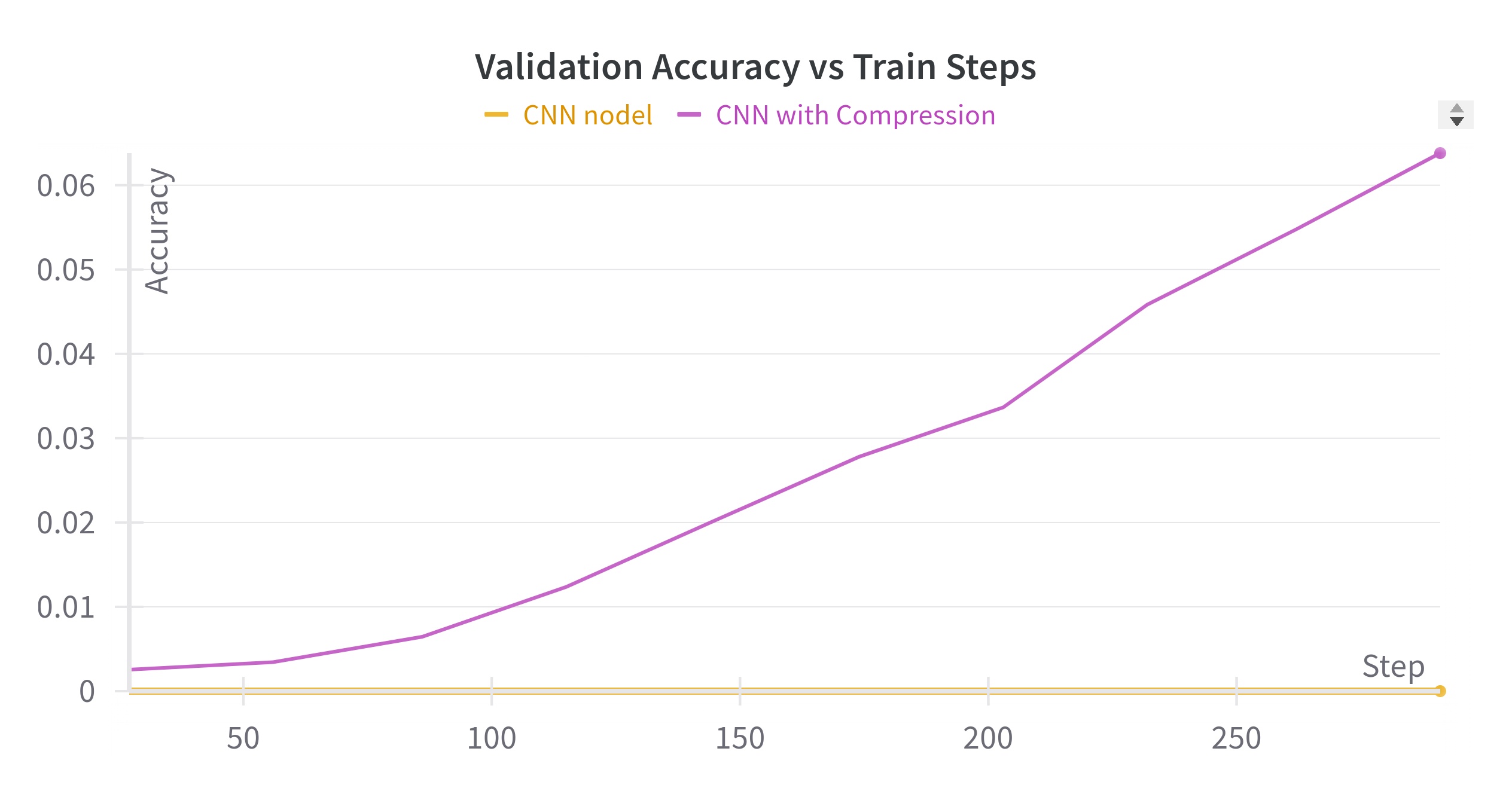}
    \caption{Validation accuracy vs Train steps for the synthetic image classification dataset.}
    \label{fig:synthetic-val-accuracy}
\end{figure}

%%%%%%%%%%%%%%%%%%%%%%%%%%%%%%%%%%%%%%%%%%%%%%%%%%%%%%%%%%%%

\section{Memory Usage}

We empirically verify that our approach significantly saves on memory and is upto 3.4x times more efficient in certain scenarios.

\paragraph{Setting}{
We monitor the memory usage during a short training epoch (100 batches) and a single validation epoch. We report the peak memory reserved by the program for training. We analyze 4 models per architecture with varying hidden sizes. All models use 8 layers and 8 heads per layer. The model of size 8.1 M uses $h=128$, the 19.3 M uses $h=256$, the 152 M uses $h=1024$, and the 659 M uses $h=2048$.
}

\paragraph{Results}{
Table~\ref{tab:memory_usage_comparison} contains the results of this experiment, "oom" stands for out of memory. We see that we achieve non-trivial memory efficiency simply by compressing the vocabulary layer. Our results show that we can double (T4) or in some cases triple (A10G) the model size and still train the model on the same GPU.
}

\begin{table}[h]
\centering
\caption{Memory Usage Comparison for GPT-2, GPT-Neo, and Our Implementation on TinyStories during a training loop. "oom" stands for out of memory.}
\begin{tabular}{l|l|ccc|r}
\toprule
\textbf{} & & \multicolumn{3}{c}{\textbf{Memory Usage} (GB)} & \textbf{} \\
\cmidrule(lr){3-5}
GPU & Model Size & GPT-2 & GPT-Neo & Ours & Efficiency \\
\midrule
\multirow{4}{*}{T4} & 8.1 M & oom & 13.0 & \textbf{3.80} & 3.4x \\
                     & 19.3 M & oom & oom & \textbf{5.20} & na \\
                     & 152 M & oom & oom & oom & na \\
                     & 659 M & oom & oom & oom & na \\
\midrule
\multirow{4}{*}{A10G} & 8.1 M & 14.0 & 13.0 & \textbf{3.80} & 3.4x \\
                       & 19.3 M & 16.0 & 18.0 & \textbf{5.20} & 3.4x \\
                       & 152 M & oom & oom & \textbf{16.6} & na \\
                       & 659 M & oom & oom & oom & na \\
\midrule
\multirow{4}{*}{L40s} & 8.1 M & 14.0 & 13.0 & \textbf{3.80} & 3.4x \\
                       & 19.3 M  & 16.0 & 18.0 & \textbf{5.20} & 3.4x \\
                       & 152 M & 28.0 & 30.0 & \textbf{16.6} & 1.8x \\
                       & 659 M & oom & oom & \textbf{39.0} & na \\
\bottomrule
\end{tabular}
\label{tab:memory_usage_comparison}
\end{table}

%%%%%%%%%%%%%%%%%%%%%%%%%%%%%%%%%%%%%%%%%%%%%%%%%%%%%%%%%%%%

\section{Computational Efficiency}

Our model significantly reduces the number of parameters required in the vocabulary layer. This not only reduces the average flops, but it also significantly speeds up the performance of the forward pass.

\paragraph{Setting}{
Similar to memory usage, we run a single short training loop and report the average throughput in terms of tokens per second. We also report the average FLOPs used per forward pass which is another strong measure of computational efficiency.
}

\paragraph{Results}{
Table~\ref{tab:throughput_flops_comparison} showcases our results for performance in terms of throughput and flops. We see a staggering 3x (8.1 M) increase in throughput and a 5x (8.1 M) decrease in FLOPs used all while maintaining performance on language modelling. We see that the improvement decreases with increase in model size as the vocabulary layer becomes less of a bottleneck.
}

\begin{table}[h]
\centering
\caption{Token Throughput and FLOPs Comparison for GPT-2, GPT-Neo, and Our Implementation on a single L40s.}
{
\begin{tabular}{l|ccc|ccc}
\toprule
& \multicolumn{3}{c}{\textbf{Throughput} (Tokens/s)} & \multicolumn{3}{c}{\textbf{FLOPs} (GFLOPs)} \\
\cmidrule(lr){2-4} \cmidrule(lr){5-7}
Model Size & GPT-2 & GPT-Neo & Ours & GPT-2 & GPT-Neo & Ours \\
\midrule
8.1 M & 108373 & 80757 & 313049 & 4.10 & 4.64 & 0.83 \\
19.3 M & 90458 & 69967 & 234268 & 9.00 & 10.0 & 3.30 \\
152 M & 32194 & 31523 & 59106 & 72.0 & 82.0 & 52.0 \\ 
\bottomrule
\end{tabular}}
\label{tab:throughput_flops_comparison}
\end{table}

%%%%%%%%%%%%%%%%%%%%%%%%%%%%%%%%%%%%%%%%%%%%%%%%%%%%%%%%%%%%

\section{Ablation Studies}

The group size is a key hyper parameter that can greatly impact the performance and efficiency of our method. We perform an ablation study varying the group size while monitoring $\mathcal{L_{\mathrm{val}}}$ and memory used.

\paragraph{Setting}{
We train a series of model on TinyStories, and for the purpose of the ablation, we train a small model with hidden size $64$, $4$ layers and $2$ heads per layer for half an epoch.
}

\paragraph{Results}{
Figure~\ref{fig:ablations} showcases the study. As discussed in Section~\ref{sec:optimality}, we see optimal memory usage when group size is equal to the square root of the vocabulary. We also see that this is the maxima in terms of val loss. This makes sense as we have the most inter-group confusion at this point, however, the performance difference isn't very large varying between $3.7$ and $3.9$ for most group sizes around $\sqrt{v}$.
}

\begin{figure}[h]
    \centering
    \includegraphics[width=1\linewidth]{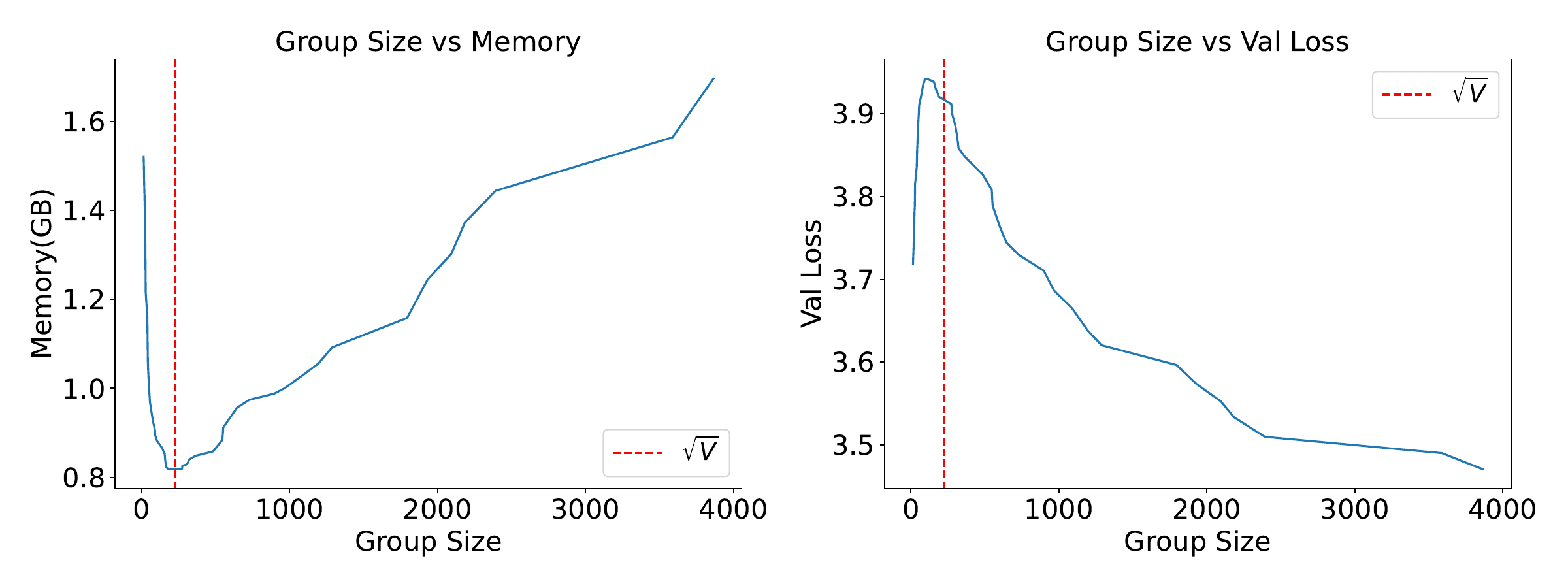}
    \caption{Ablation studies}
    \label{fig:ablations}
\end{figure}

%%%%%%%%%%%%%%%%%%%%%%%%%%%%%%%%%%%%%%%%%%%%%%%%%%%%%%%%%%%%

\section{Limitations \& Conclusion}

While our method improves memory efficiency and computational performance, several limitations remain. Despite surpassing GPT-2 and GPT-Neo models, we were unable to empirically compare with other vocabulary compression techniques due to compute constraints. Additionally, our method shows sensitivity to the group size hyperparameter, as noted in our ablation study. Limited training epochs also impacts results, but we believe they are representative of low-compute settings.

In this work, we introduced a novel approach to compressing the final vocabulary layer, achieving both memory efficiency and high performance. We hope this work contributes to low-compute machine learning and demonstrates how simple optimisations can highly effective and practical.

% In this work, we introduced a novel approach to compressing the final vocabulary layer that achieves both optimal memory efficiency and high performance. We extensively validated our model on the TinyStories dataset, demonstrating comparable performance while achieving significant memory reductions and improved throughput. We hope this work furthers the field of low compute machine learning and illustrates how simple optimisations can greatly enhance the effectiveness of low-resource environments.

%%%%%%%%%%%%%%%%%%%%%%%%%%%%%%%%%%%%%%%%%%%%%%%%%%%%%%%%%%%%

\section{Acknowledgement}

The authors of this paper would like to thank Shashwat Singh, Abhinav S. Menon, and Vamshi Krishna Bonagiri for their help and thoughtful comments. We also thank Lightning.ai for free access to quality GPUs for our experiments.

%%%%%%%%%%%%%%%%%%%%%%%%%%%%%%%%%%%%%%%%%%%%%%%%%%%%%%%%%%%%

\bibliographystyle{apalike} % Choose a bibliography style
\bibliography{references}

% \section*{References}

% {
% \small

% [1] Alexander, J.A.\ \& Mozer, M.C.\ (1995) Template-based algorithms for
% connectionist rule extraction. In G.\ Tesauro, D.S.\ Touretzky and T.K.\ Leen
% (eds.), {\it Advances in Neural Information Processing Systems 7},
% pp.\ 609--616. Cambridge, MA: MIT Press.

% }

%%%%%%%%%%%%%%%%%%%%%%%%%%%%%%%%%%%%%%%%%%%%%%%%%%%%%%%%%%%%

\clearpage

\appendix

\section{Psuedocode}\label{app:psuedo}

An alternative that we initially explored involved storing a unique linear tensor for each group (Implementation~\ref{fig:slow_implementation}) instead of a shared linear with a scale and shift (Implementation~\ref{fig:fast_implementation}). This forced us to loop over the groups if we wished to keep memory usage low. Even with a clever implementation using masks, this approach was significantly slower and was replaced with the scale and shift tensors.

\begin{figure}[h]
    \centering
    \begin{minipage}{\textwidth}
    \captionsetup{type=listing}  % Allow captioning for code blocks
    \begin{lstlisting}[language=Python][frame=lines, framesep=2mm, fontsize=\small, linenos, numbersep=5pt]{python}
def apply_linear(self, h, groups):
    batch_size, sequence_length, hidden_size = h.shape
    output = torch.zeros(batch_size, sequence_length,
        self.group_size, device=h.device)
    
    h_flat = h.view(-1, hidden_size)
    output_flat = output.view(-1, self.group_size)
    groups_flat = groups.view(-1)
    
    for i in range(self.num_groups):
        mask = (groups_flat == i)
        if mask.any():
            group_input = h_flat[mask]
            group_output = self.linears[i](group_input)
            output_flat[mask] = group_output
            
    return output_flat.view(batch_size, sequence_length, self.group_size)
    \end{lstlisting}
    \end{minipage}
    \caption{Slow implementation which requires looping over the groups}
    \label{fig:slow_implementation}
\end{figure}

\begin{figure}[h]
    \centering
    \begin{minipage}{\textwidth}
    \captionsetup{type=listing}  % Allow captioning for code blocks
    \begin{lstlisting}[language=Python][frame=lines, framesep=2mm, fontsize=\small, linenos, numbersep=5pt]{python}
 def apply_linear(self, h, groups):
    shared_output = self.shared_linear(h)

    groups_flat = groups.view(-1)
    shared_output_flat = shared_output.view(-1, self.group_size)

    scale = self.scale(groups_flat)
    shift = self.shift(groups_flat)

    modulated_output_flat = shared_output_flat * scale + shift
    modulated_output = modulated_output_flat.view_as(shared_output)

    return modulated_output
    \end{lstlisting}
    \end{minipage}
    \caption{Fast implementation applying the scale and shift transformation.}
    \label{fig:fast_implementation}
\end{figure}

% \begin{figure}[h]
%     \centering
%     \includegraphics[width=1\linewidth]{figures/tiny-storiestrain-loss.png}
%     \caption{Train loss curves for our models on TinyStories}
%     \label{tinystories-val-loss-curves}
% \end{figure}

\section{Multi Class Classification} \label{app:multiclass}

\subsection{Dataset}\label{synthetic-dataset}

The Dataset comprised of 184k classes each containing about 10 images. Each class comprised of the following attributes.

\begin{table}[h]
    \centering
    \begin{tabular}{ll}
        \toprule
        \textbf{Attribute} & \textbf{Options} \\
        \midrule
        Shapes & Circle, Square, Triangle, Pentagon, Hexagon, Octagon, Star, Cross \\
        Patterns & Solid, Striped, Dotted, Checkered \\
        Rotations & 0°, 60°, 120°, 180°, 240°, 300° \\
        Colors & Red, Green, Blue, Yellow, Purple, Orange, Pink, Brown, Cyan, Magenta, Lime, Teal \\
        Sizes & Tiny, Small, Medium, Large, Huge \\
        Textures & None, Noise \\
        Opacities & 0.25, 0.5, 0.75, 1.0 \\
        Border Styles & Solid, Dashed \\
        \bottomrule
    \end{tabular}
    \caption{Attributes and Options}
\end{table}

One attribute from each class was randomly chosen to generate about 2M images. Every possible class was mapped to an index and stored along with the images.

\subsection{Model Architecture}

We used a 3 layer CNN with Kernel Size = 3 and padding = 1. Channels were taken from 3 to 32 in the first layer and doubled in each subsequent layer. The output was flattened and subsequently encoded into a 512-dimensional embedding through a fully connected layer along with Relu non linearity.

The baseline model linearly mapped this to the vocab size, whereas our method mapped it into one of the ${\sqrt{V}}$ groups and subsequently mapped to a particular token.

\subsection{Training}

The training was conducted for 10 epochs with a batch size of 64 and gradient accumulation over 4 batches. Learning rate was set to 0.001 and Adam optimizer was used to update the weights. Cross Entropy Loss was used to train and evaluate the model's performance. Accuracy was recorded at the end of each epoch over the Validation set.

\subsection{Train Plots}

Surprisingly, the group accuracy is quite high, this suggests that groups might be easier to learn than the complete task and sheds light on why our approach performs so well in practice despite its loss in theoretical expressivity.

\begin{figure}[h]
    \centering
    \includegraphics[width=0.5\linewidth]{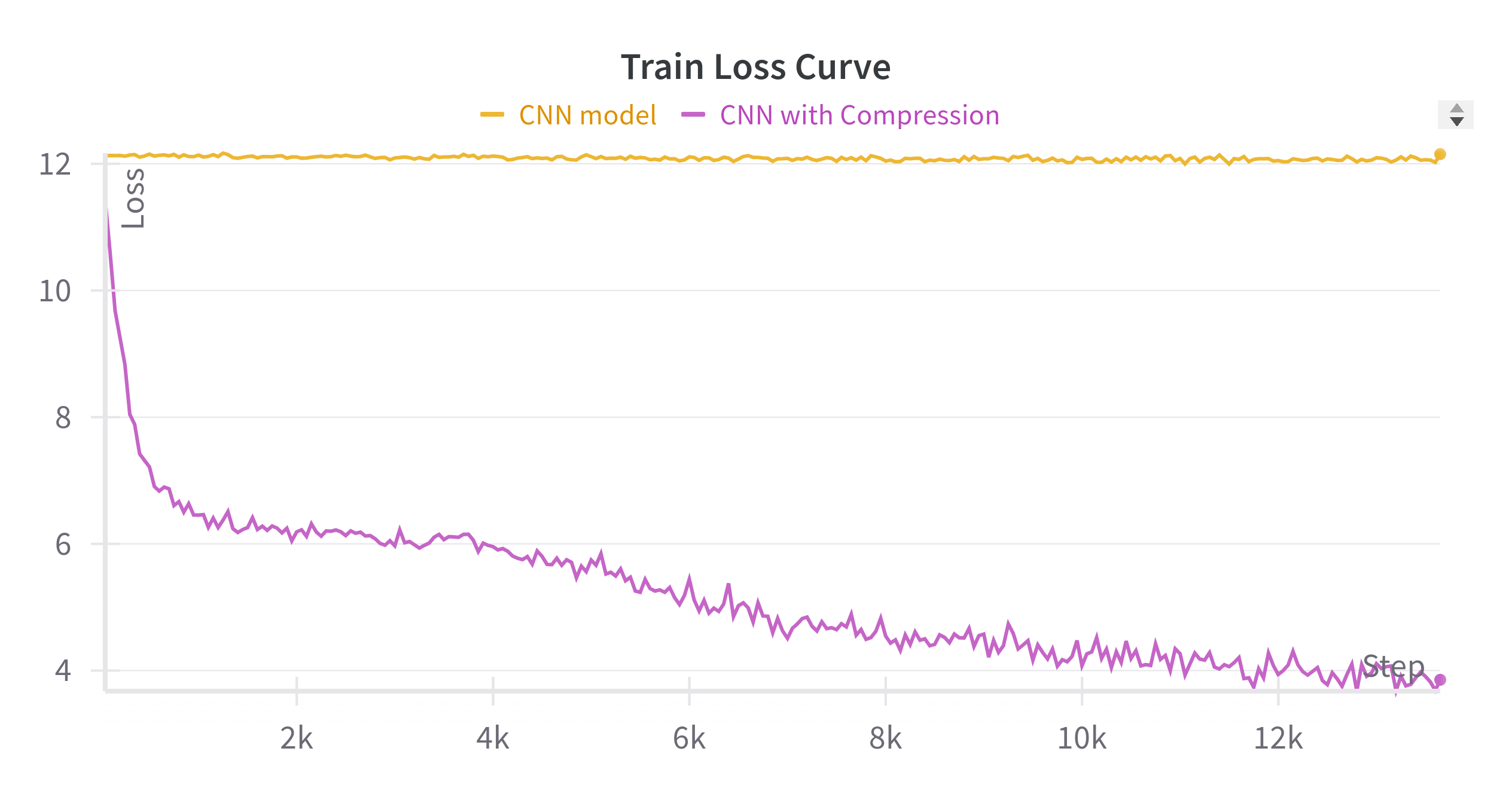}
    \caption{Train loss curve comparison on multi-class classification}
\end{figure}

\label{app:groupaccuracyimg}

\begin{figure}[h]
    \centering
    \includegraphics[width=0.5\linewidth]{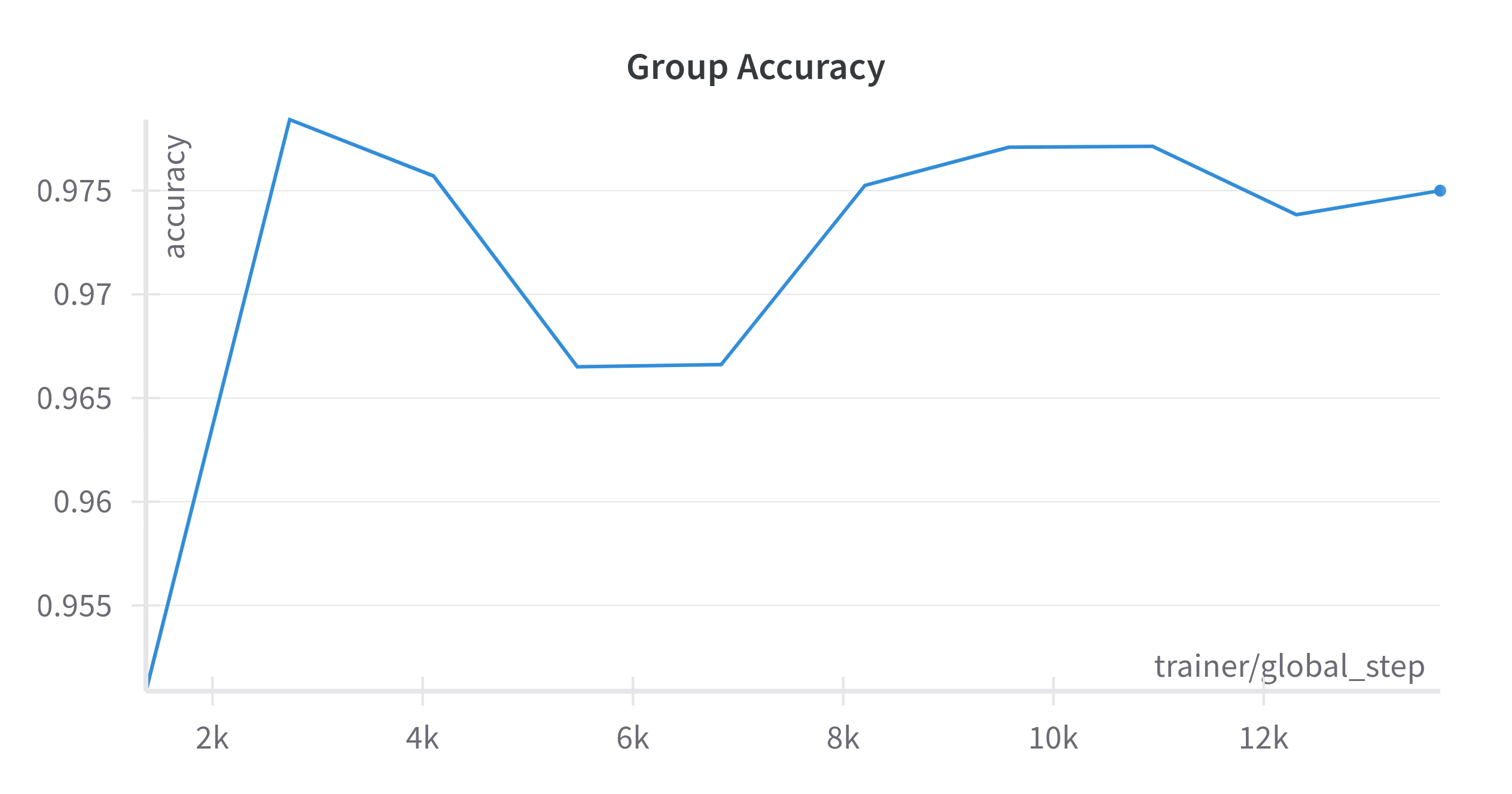}
    \caption{Group prediction accuracy of our model. We see that out model takes advantage of the fact that smaller groups are easier to learn than the whole dataset}
\end{figure}

%%%%%%%%%%%%%%%%%%%%%%%%%%%%%%%%%%%%%%%%%%%%%%%%%%%%%%%%%%%%

\end{document}